\documentclass[conference]{IEEEtran}
\IEEEoverridecommandlockouts
\usepackage{cite}
\usepackage{amsmath,amssymb,amsfonts}
\usepackage{algorithmic}
\usepackage{graphicx}
\usepackage{textcomp}
\usepackage{xcolor}
\usepackage{url}
\usepackage[linkcolor=black]{hyperref}
\def\BibTeX{{\rm B\kern-.05em{\sc i\kern-.025em b}\kern-.08em
    T\kern-.1667em\lower.7ex\hbox{E}\kern-.125emX}}
\begin{document}
\title{Unknown Object Segmentation through Domain Adaptation\\

}
\author{\IEEEauthorblockN{Yiting Chen}
\IEEEauthorblockA{\textit{School of Power and Mechanical Engineering} \\
\textit{Wuhan University}\\
Wuhan, China \\
chenyiting@whu.edu.cn}
\\
\IEEEauthorblockN{Chenguang Yang}
\IEEEauthorblockA{\textit{Bristol Robotics Laboratory} \\
\textit{University of the West of England}\\
Bristol, UNITED KINGDOM \\
chenguang.yang@brl.ac.uk}
\and
\IEEEauthorblockN{Miao Li}
\IEEEauthorblockA{\textit{School of Power and Mechanical Engineering} \\
\textit{Wuhan University}\\
Wuhan, China \\
limiao712@gmail.com}

}

\maketitle

\begin{abstract}
The ability to segment unknown objects in cluttered scenes has a profound impact on robot grasping. The rise of deep learning has greatly transformed the pipeline of robotic grasping from model-based approach to data-driven stream, which generally requires a large scale of grasping data either collected in simulation or from real-world examples. In this paper, we proposed a sim-to-real framework to transfer the object segmentation model learned in simulation to the real-world. First, data samples are collected in simulation, including RGB, 6D pose, and point cloud. Second, we also present a GAN-based unknown object segmentation method through domain adaptation, which consists of an image translation module and an image segmentation module. The image translation module is used to shorten the reality gap and the segmentation module is responsible for the segmentation mask generation. We used the above method to perform segmentation experiments on unknown objects in a bin-picking scenario. Finally, the experimental result shows that the segmentation model learned in simulation can be used for real-world data segmentation.
\end{abstract}

\section{Introduction}
Unknown objects instance segmentation, namely the ability to determine which object each pixel pertains to in cluttered scenes, can improve the ability of robot's grasping. Some recent works can generate grasping policies from various datasets of point cloud\cite{li2016dexterous}, images\cite{zhang2020robotic}, grasping attempts\cite{levine2018learning}. While these approaches are superb at analyzing complex object stacking scenes and generating grasping strategies, but they do not distinguish between the objects they are grasping. Though, these methods can be extended to plan grasps for a 
specific target object by using an object mask to restrict the grasp planning.
\begin{figure}[htbp]
	\center
	\includegraphics[scale=0.5]{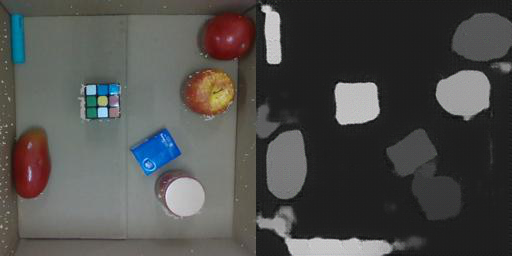}
\caption{Unknown Object Instance Segmentation}
\end{figure}
With the help of semantic information, some recent approaches perform really excellent on image segmentation, such as U-net\cite{10.1007/978-3-319-24574-4_28}, R-Mask CNN\cite{He_2017_ICCV}, these methods require time-consuming human labeling to generate training data, and existing datasets consist of RGB images don't seem to have the types of cluttered scenes mostly encountered in the bin-picking scenario. Collecting additional datasets for new scenario requires time-consuming manual labeling. Without the prior knowledge of semantic information, it is difficult to segment unknown objects due to the luster, sensor noise, and other distractions. Therefore, pixel-wise object segmentation is barely used in robotic grasping.

To address these issues, we directly generate a large amount of labeled data in the simulator, and use domain adaptation method to directly apply the network trained in the simulation to the real world examples. Two generative adversarial networks are combined to achieve domain adaptation and perform category-agnostic instance segmentation on noisy real RGB images without training on real hand-labeled data. 
In order to learn the instance segmentation network from simulation, we propose to train on a large dataset of synthetic RGB images with domain randomization over a various set of 3D objects, textures, lighting sources. 
  
  This paper contributes:
\begin{enumerate}
\item A method for quickly generating a synthetic dataset of RGB, depth images, segmentation masks, point cloud, and 6D position using domain randomization with PyBullet\cite{coumans2016pybullet}.
 
\item Double-bidirectional segmentation network consists of two independent module, designed to perform unknown object instance segmentation on real RGB images, trained on synthetic data.
  
\end{enumerate}

\section{Related Work}
\subsection{Domain Adaptation}
Domain adaptation is a widely used technique in robotics and computer vision. During robot learning, one of the main difficulties comes from the burden of data collection with physical system. It is usually time consuming and costly. In \cite{andrychowicz2020learning}, the authors proposed an approach to learn the  motion policy in simulated environment, which is transferred to real robot using domain adaptation. For more complex tasks, like dexterous manipulation, the authors boost the learning process with simulation, in which many key physical parameters (friction, mass, appearance, etc) are randomized. 

Domain adaptation is not only used in the transfer of dynamic strategies, but also in computer vision. Many methods perform feature-level and pixel-level adaptation simultaneous in many vision tasks, e.g., image classification, semantic segmentation, and depth estimation\cite{cheng2018depth}. Generative adversarial network (GAN) is a class of machine learning frameworks using deep learning method. CycleGAN is one of the most amazing ideas in style transfer, which translates images from one domain to another with pixel-level adaptation via image-to-image translation networks\cite{zhu2017unpaired}. In this work, we take advantage of CycleGAN to achieve domain adaptation, which transfers the real-world noisy RGB images into the synthetic non-photorealistic images.

\subsection{Image Segmentation}
Image segmentation is a very critical field in image processing, which plays an indispensable role in robotic perception. Semantic segmentation performs pixel-level labeling with a set of object categories for all image pixels. Instance segmentation further expands the scope of semantic segmentation by detecting and depicting each object of interest in the image. With the inspiration of generative adversarial networks, such as CycleGAN and Pix2pix, researchers even use conditional generative adversarial networks to generate the semantic or instance segmentation mask\cite{kniaz2018conditional}. Most of the segmentation network is based on detecting the object. Without semantic information, the segmentation task is much harder. Recent approaches also perform good methods on category-agnostic object instance segmentation based on depth\cite{danielczuk2019segmenting} images or RGB-D\cite{xiang2020learning} information. In this work, we take advantage of the Pix2pix\cite{Isola_2017_CVPR} network to do the segmentation mask generation, which doesn't need any hand-labeled semantic information.
\section{Problem Statement}
We consider the problem of unknown object instance segmentation as finding subsets of pixels corresponding to unique unknown objects in a single RGB image.
We make the following definitions to define the unknown object instance segmentation task:
\begin{enumerate}
\item \textit{Reality:} Let $\textit{D}_{R}$ present the information carried by RGB images dataset from reality, $\textit{D}_{R} =\{\overbrace{x_{r1}+x_{r2}+\cdots+x_{rn}}^{n},N_{r},C_{r}\}$. In this set, $x_{ri}$ is defined by the  geometric feature for each image, while $N_{r}$ and $C_{r}$ is defined by the noise and camera parameters.
\item \textit{Simulation:} Let $\textit{D}_{S}$ present the information carried by RGB images dataset from simulation, $\textit{D}_{S} =\{\overbrace{x_{s1}+x_{s2}+\cdots+x_{sm}}^{m},\overbrace{y_{s1},y_{s2}+\cdots+y_{sm}}^{m},N_{s},C_{s}\}$. In this set, $x_{si}$ is defined by the  geometric feature for each image, $y_{si}$ is defined by the ground truth segmentation mask for each image, while $N_{s}$ and $C_{s}$ is defined by the noise and camera parameters. Each geometric feature $x_{si}$ corresponds to a segmentation mask $y_{si}$.
\item \textit{Segmentation $G(x)$:} We use $x_{si}, N_{s}, C_{s}\in D_{S}$ as the input and $y_{si}\in D_{S}$ as the output to train the segmentation mask generation network $G(x_{si},N_{s},C_{s}\in D_{s}) = y_{si}\in D_{s}$.

\item \textit{Adaptation $f(x)$:} $D_{R}$ will be transfered into $D_{R-S}$, while $x_{r-si} \in D_{R-S}$ will be consistent with $x_{ri} \in D_{R}$, and $N_{r-s},C_{r-s}\in D_{R-S}$ will be as close as possible to $N_{s},C_{s}\in D_{S}$, $f(D_{R}) = D_{R-S}$.

\end{enumerate}
\section{Proposed Method}
According to the our definition, every real image contains geometric feature which corresponds to a segmentation mask $\mathcal{M_{R}} = \{(\mathcal{M}_{i}:\mathcal{M}_{i}\neq\emptyset)\}$ . We generate unknown objects segmentation mask from reality as follow:
$$f(x_{ri}, N_{r}, C_{r} \in D_{R}) = x_{ri}, N_{r-s}, C_{r-s}\in D_{R-S}$$
$$G(x):{x_{si},N_{s},C_{s}}\in D_{R} \to y_{si}\in D_{R}   $$
$$ \mathcal{M}_{i} = G[f(x_{ri}, N_{r}, C_{r} \in D_{R})]  $$
\begin{figure*}[htbp]
    \centering
    \includegraphics[scale=0.5]{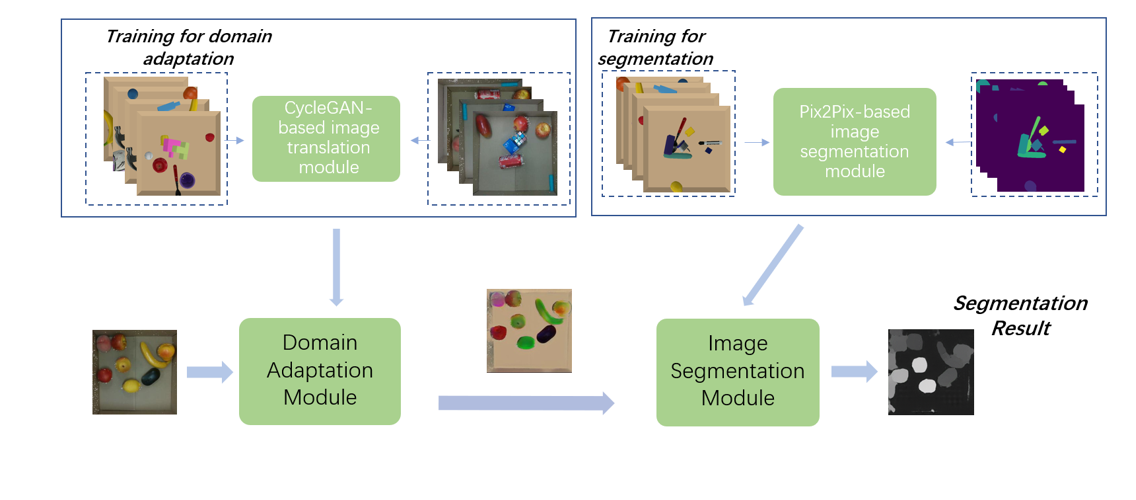}
    \caption{The schematic flow chart of our method. The first module is image translation, which attempts to translate the data from reality domain $\textit{D}_{R}$ into the domain $\textit{D}_{R-S}$ which is as close to the simulation domain  $\textit{D}_{S}$ as possible. The second module is image segmentation, which generates instance segmentation mask by a end-to-end approach. }
    \label{flow chart}
    \end{figure*} 
\subsection{Synthetic Data Generation}
We choose PyBullet \cite{coumans2016pybullet} as our simulator, which provides real-time collision detection and multi-physics simulation. Part of our objects are selected from the YCB dataset \cite{calli2015benchmarking}, which provides nearly a hundred kinds of texture-mapped 3D mesh models, and the rest are randomly generated object with irregular shape and color provided by PyBullet, which are mainly composed of cubes, triangular pyramids and spheres of different sizes.

\begin{figure}[htbp]
    \centering
    \includegraphics[scale=0.5]{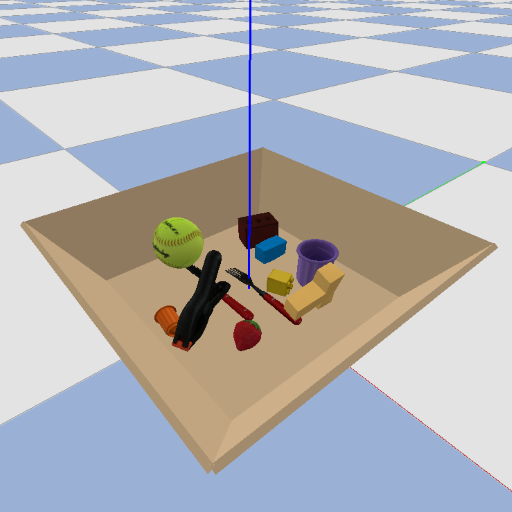}
    \caption{We choose pybullet as our simulator: 8 different objects randomly selected from our data base, and fall into the tray box to imitate the bin-picking scenario. }
    \label{bin-picking}
    \end{figure}

\begin{figure}[htbp]
	\centering
	\includegraphics[scale=0.3]{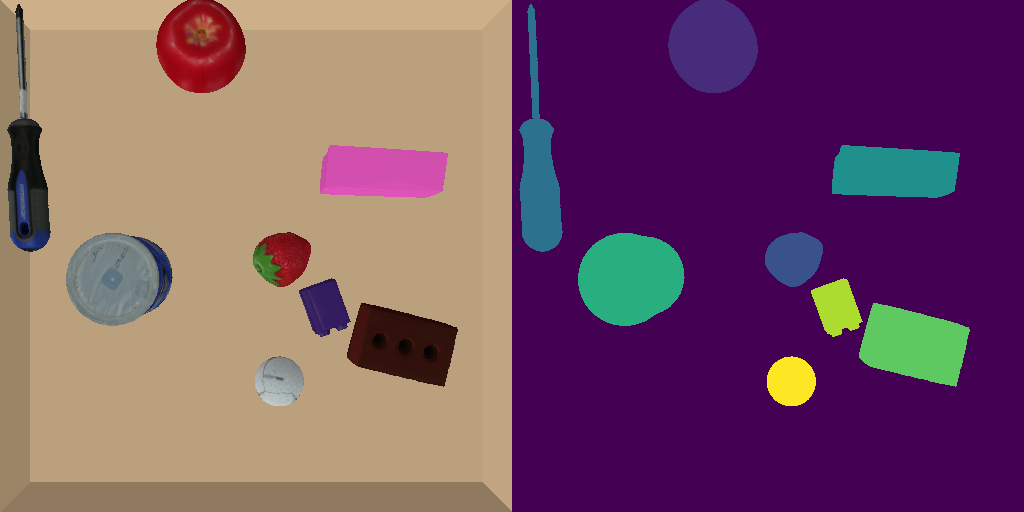}
\caption{Because of the powerful built-in functions of pybullet, we could collect paired pictures easily. A pair of synthetic RGB and segmentation mask from the bin-picking scenario, which both contains 256*256 pixels.}
\label{data1}
\end{figure}

The virtual environment we designed is to place an empty tray box in the middle of a blank plane and the camera 0.7 meters above the tray box to imitate a bin-picking scenario. We set a blank space of 0.4*0.4*0.45 cubic meters and make it 0.05 meters right above the tray box. Each time we randomly selected $n$, $n\in[7,8,9,10,11]$ different kinds of models to appear from a random position above the box, every single object’s x,y,z parameters were generated randomly within the size of the blank space. 
Once we turned on gravity, the objects would naturally fall into the tray box. Due to the collision, the poses of each object were naturally randomly generated, so that the stacking states of objects were very similar to the real-world situation. Figure \ref{bin-picking} shows the situation of our simulation. 
For each falling case, the lighting of the scene comes from a point light that will constantly change its angle, which means we could obtain nearly every lighting situation that is possible in the real world. We collect the object masks $y_{si}\in D_{s}$ , which is determined by the geometric feature $x_{si}\in D_{s}$ carried by the synthetic RGB image. Figure \ref{data1} shows a pair of our synthetic data. \url{https://github.com/ChenEating716/CYT_Dataset} shows the code for data generation.
  
\begin{figure}[htbp]
    \centering
    \includegraphics[scale=0.3]{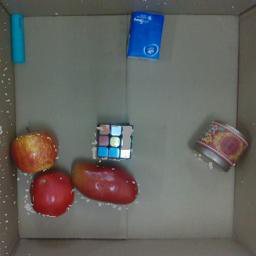}
    \includegraphics[scale=0.3]{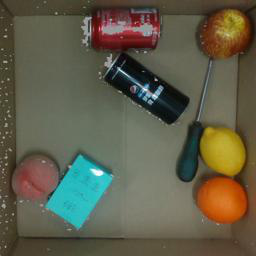}
    \includegraphics[scale=0.3]{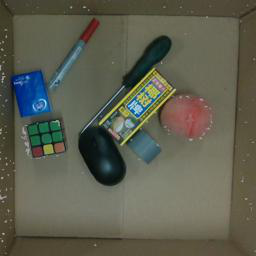}
    \includegraphics[scale=0.3]{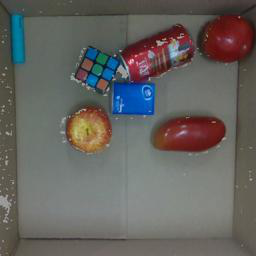}
    \includegraphics[scale=0.3]{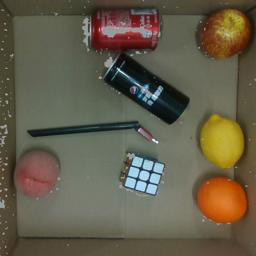}
    \includegraphics[scale=0.3]{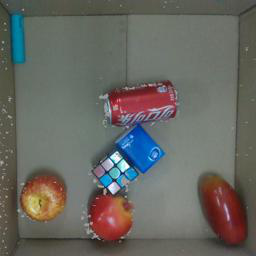}
    \caption{This is the real bin-picking scenario which needs to be segmented. Different objects randomly selected from our office, and fall into a paper box.  }
    \label{realdata}
    \end{figure}

In order to transfer the segmentation network effectively, we collect a small number of pictures from a real bin-picking scenario. The collecting process is similar to the simulation because we need to let the domain adaptation focus on the $N_{r}, C_{r}\in D_{R}$ instead of the $x_{ri}\in D_{R}$. We put an Intel D435 camera above the bin and throw daily objects in it. The object position and occlusion are generated randomly, while the $x_{ri}, N_{r}, C_{r}\in D_{R}$ are easy to collect. Figure \ref{realdata} shows the real scenes of the bin-picking.

\subsection{Image Translation Module}
In order to allow our synthetic data to better act on reality without losing its meaning due to the reality gap, the function of the first module in our method is domain adaptation $f(x)$. In order to maximize the function of the segmentation module, we have to transfer the real data $D_{r}$ to the domain $D_{r-s}$ similar to simulation, which means distribution $\mathcal{P}(D_{r-s})\approx\mathcal{P}(D_{s})$. At the same time, we have to ensure that during the translation process, the geometric information carried by the RGB image itself $x_{ri}\in D_{R}$ will not change due to the change of the domain. Based on the CycleGAN network structure, we achieve this goal.

\begin{figure}[htbp]
    \centering
    \includegraphics[scale=0.3]{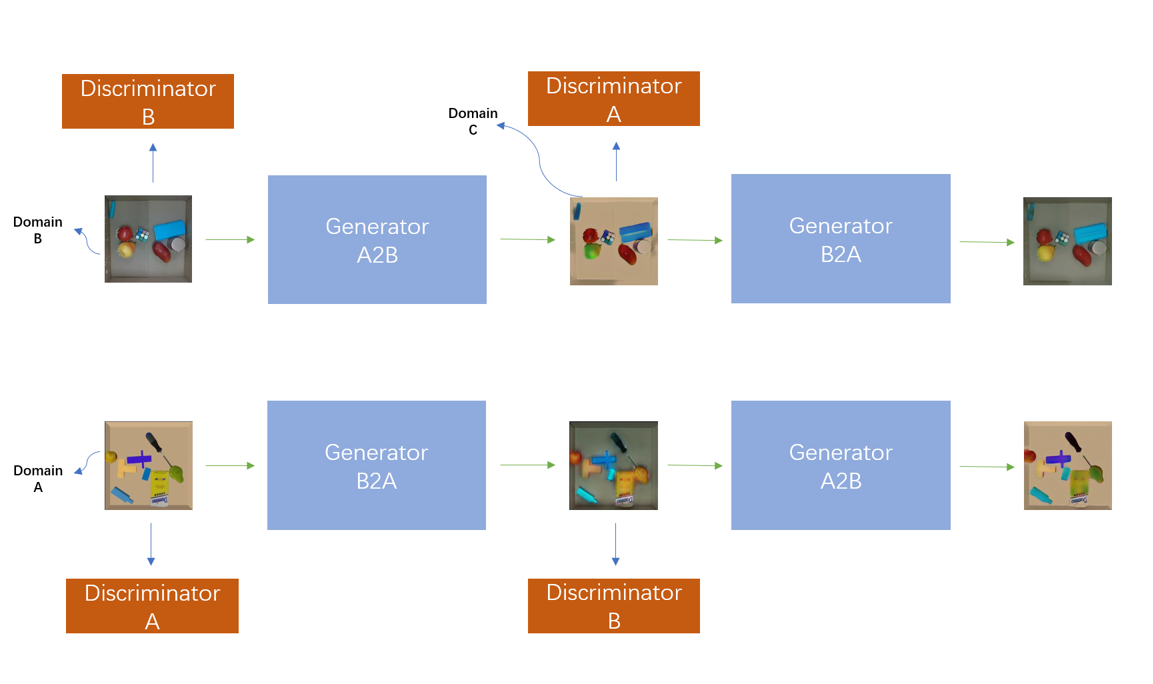}
    \caption{Basically a simplified view of CycleGAN architecture, which demonstrates how the domain adaptation process is achieved. With the help of CycleGAN, we successfully transform the real images into simulation style.  }
    \label{cyclegan}
    \end{figure}

\begin{figure}[htbp]
	\center
	\includegraphics[scale=0.4]{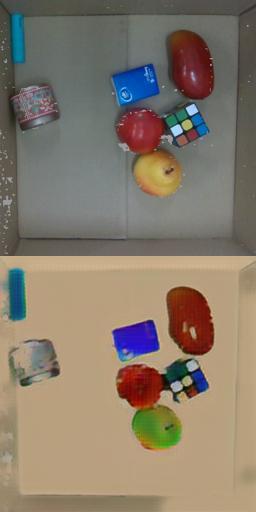}
	\includegraphics[scale=0.4]{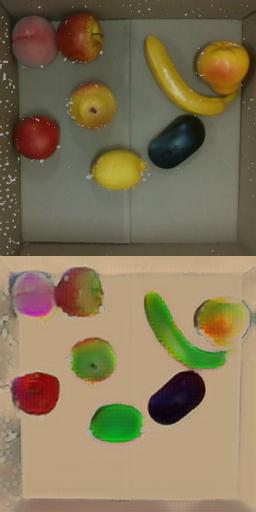}	
	\includegraphics[scale=0.4]{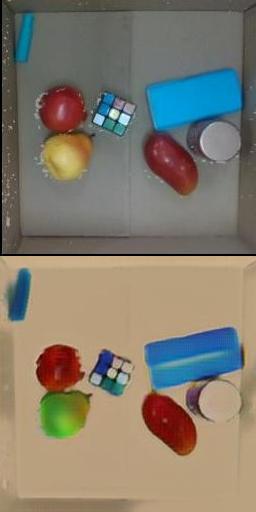}
\caption{Noisy pictures from reality is translated into "simulation style", which maintains its original geometric information. The first row represents the noisy pictures from reality and the second row represents their "simulation style".  }
\label{real2sim}
\end{figure}

Because the synthetic images from the simulator carry much less noise and other negative distractions, we decide to make the distribution of data from reality closer to simulation. CycleGAN is consists of two generators $\mathcal{G}_{A2B}$,$\mathcal{G}_{B2A}$, and two discriminators $\mathcal{D_{A}}$,$\mathcal{D_{B}}$.  First, we label the domain which presents the distribution $\mathcal{P}(D_{s})$ of simulated data as $Domain A$, and the domain which presents the distribution $\mathcal{P}(D_{r})$ of real data as $Domain B$. Our purpose is to transfer the data from $Domain B$ to $DomainC$, which has a similar distribution $\mathcal{P}(D_{r-s})$ to $Domain A$, while keeping the feature information as consistent as possible. $\mathcal{G}_{B2A}$ is mainly responsible for transferring the data into $Domain C$, while $\mathcal{G}_{A2B}$ ensures that the geometric information is consistent by comparing the original image and the reconstructed image from $Domain C$ to $Domain B$. $\mathcal{D_{A}}$ is used to evaluate the distance between $Domain A$ and $Domain C$, and reduce the difference by iterative training. Figure \ref{cyclegan} shows the structure of the image translation module.    Figure \ref{real2sim} shows the whole translation process.

\subsection{Image Segmentation Module}
Even though there are many segmentation networks that perform well in various segmentation environments, it is still not common to segment unknown objects based on RGB images. In this work, we are different from the traditional segmentation network, which is based on detection first and then segmentation. We understand the process of segmentation mask acquisition as an end-to-end generation work, the input is the RGB image and the output is the segmentation mask according to its geometric features.

To adapt the Pix2pix network to perform unknown object instance segmentation on RGB images, we made several modifications:
\begin{enumerate}
\item We transfer the RGB images into grayscale images to meet the need of sobel operator.
\item We use Sobel operator to do the image pre-processing. Even the training data come from simulation, there is still some inevitable distractions for object segmentation, such as luster, colorful texture e.g. The Sobel operator is effective at smoothing the noise and providing more accurate edge direction information, which is extremely beneficial to the segmentation mask generation, especially in the situation that we do not pay any attention to the semantic information.

\item We transfer the output image also into grayscale. For the segmentation mask generation, color is a kind of redundant information in a sense. The grayscale image is enough to carry the segmentation information.
    \end{enumerate}

Our implementation of segmentation mask generation is mostly based on the original paper\cite{Isola_2017_CVPR}. The modifications we made are listed above and the network is trained on the dataset which is generated by our method above. 
\section{Experimental Results}
\subsection{Segmentation on Synthetic Data}
\begin{figure}[htbp]
    \centering
	\includegraphics[scale=0.4]{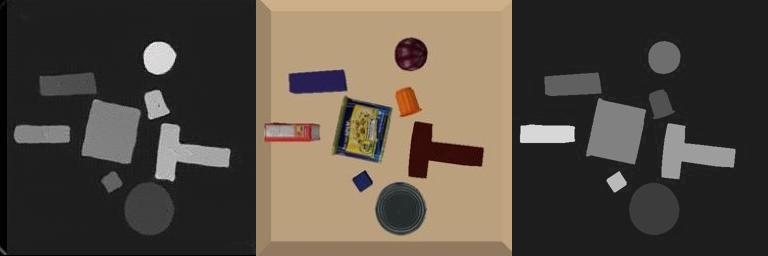}	
	\includegraphics[scale=0.4]{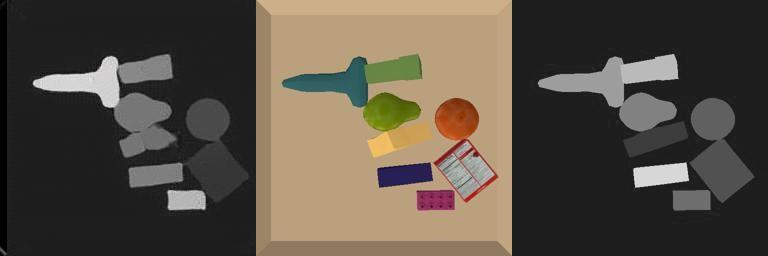}
    \caption{From left to right is segmentation mask generated by our segmentation module, synthetic image and the ground truth segmentation mask.}
    \label{sim2seg}
    \end{figure}

The image segmentation module of our method is completely trained by synthetic data. Due to the convenience of generating numerous well-labeled synthetic RGB images, the segmentation function should perform well on our synthetic images. And the result in Figure \ref{sim2seg} turns out that the segmentation mask generated by our segmentation module is very close to the ground truth segmentation mask.
\subsection{Segmentation on Real Data}
\begin{figure}[htbp] 
	\center
	\includegraphics[scale=0.15]{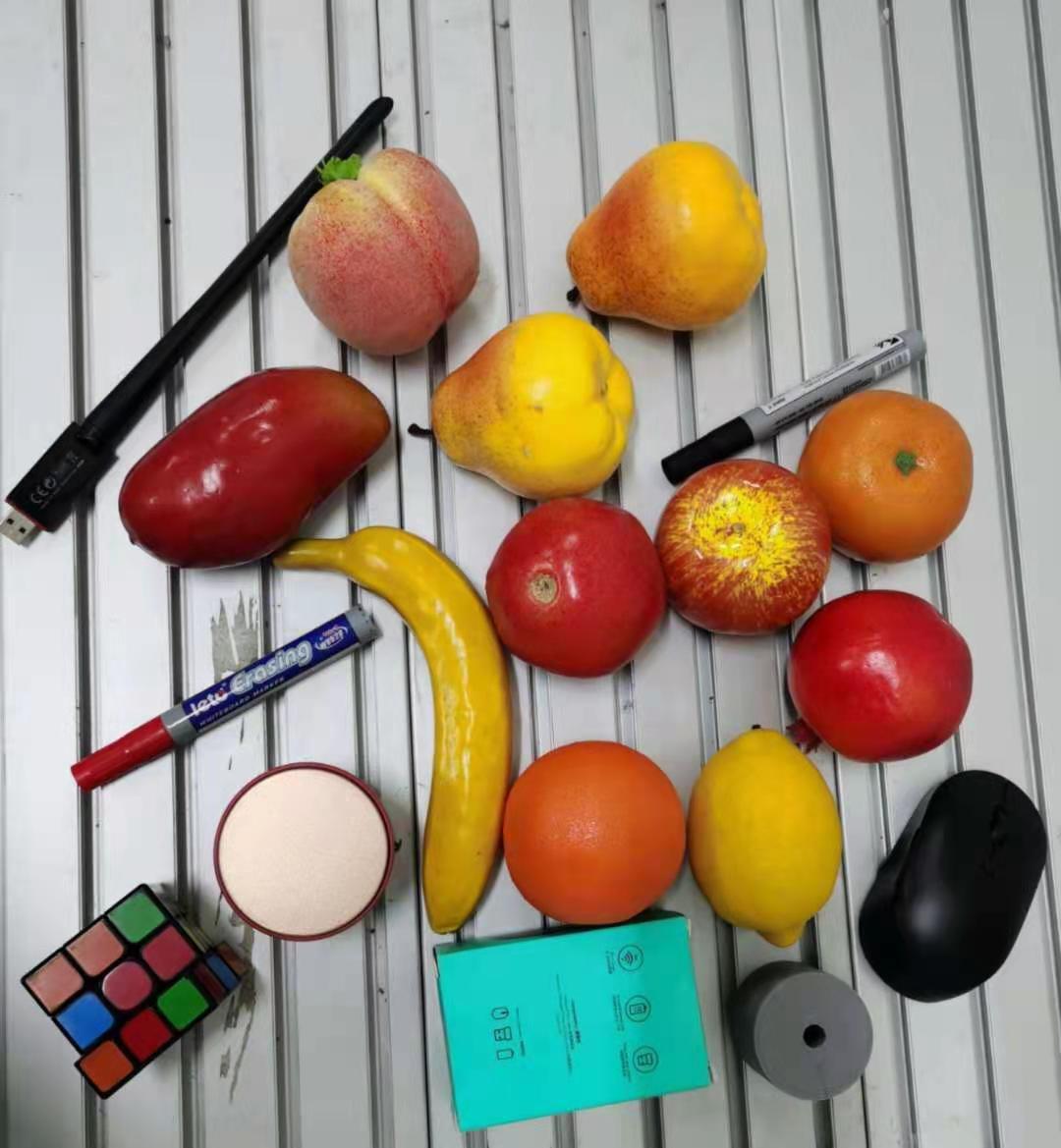}
\caption{Part of the unknown objects we used to test our method, which are basically common items in daily life. }
\label{realones}
\end{figure}

\begin{figure}[htbp] 
	\center
	\includegraphics[scale=0.4]{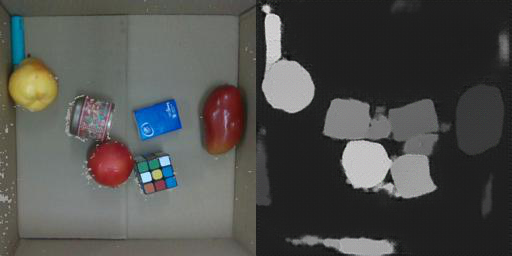}
	\includegraphics[scale=0.4]{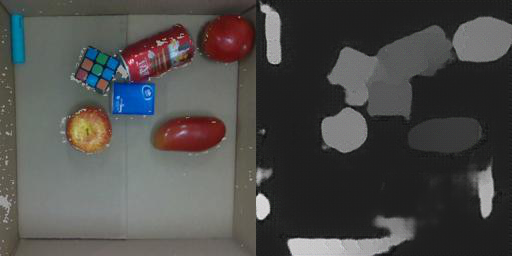}
	\includegraphics[scale=0.3]{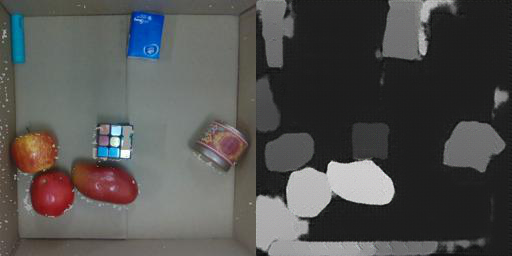}
\caption{Left row represents the noisy pictures of unknown objects from reality and right row represents their corresponding segmentation results.  }
\label{performance}
\end{figure}

One of the main purposes of our work is to prove that the feasibility of using the data generated by simulation to perform unknown object segmentation in reality exists, so testing the segmentation on synthetic data is only step one. To evaluate the segmentation ability on real images, we ran our methods on our noisy RGB images collected from reality from Realsense D435 camera. Figure \ref{realones} shows part of the objects from reality we used to test our method, which are basically common items in daily life. Five hundreds pictures were collected, all real test images were resized into 256 by 256 pixels. Figure \ref{performance} shows part of our results.
\subsection{Comparison}
Our method consists of an image translation module and a segmentation module. Because of the characteristics of the adversarial generation network, there will be some noises that are difficult to distinguish with the naked eye in the generated images. Therefore, before evaluating the effect of image segmentation, we use algorithms such as watershed to process it to achieve the effect of denoising. In order to independently estimate the effect of each module and the whole method, we use mPA(mean Pixel Accuracy) and mIoU(mean Intersection over Union) to evaluate the performance of the segmentation result, which reports the average percent of pixels in the image which were correctly classified for each object and the average ratio of the intersection and union of the predicted value and the true value for each object. Table \ref{comparision1} and Figure \ref{performance} show the performance of our method on synthetic images.
Table \ref{comparision} and Figure \ref{comparison} show the comparison of performance between the method with and without the image translation module. And   Result shows that the image translation module greatly enhances the model on weakening the distraction of interference information and shorten the gap between simulation and reality.
\begin{figure}[htbp] 
	\center
	\includegraphics[width=0.4\linewidth]{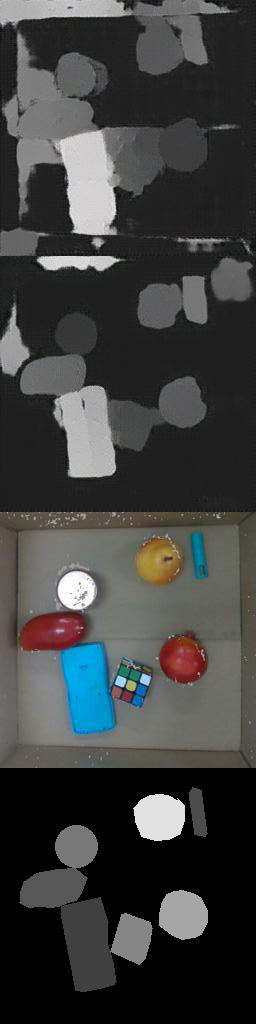}
	\includegraphics[width=0.4\linewidth]{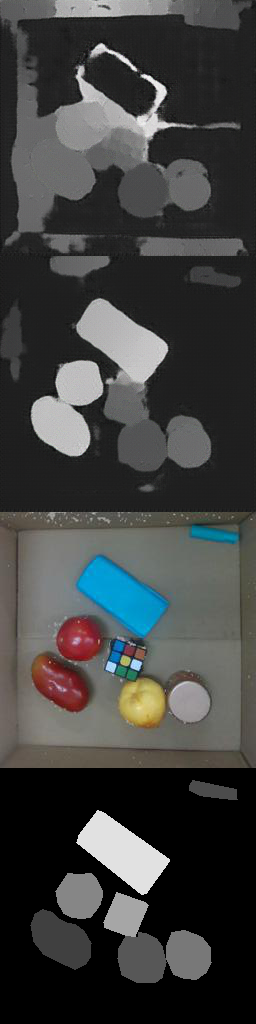}
\caption{From top to bottom, the first one is segmentation mask without image translation module(domain adaptation), the second one is the segmentation mask by our method, the third one is noisy picture from reality and the last one is the ground truth segmentation mask. }
\label{comparison}
\end{figure}
\begin{table}
\label{comparison}
\centering
\caption{Mean Pixel Accuracy and mean Intersection of Union on the segmentation result on real images with and without domain adaptation. The domain adaptation process greatly improved the segmentation ability in real world.}
\label{comparision}   
\begin{tabular}{lll}
\hline\noalign{\smallskip}
Method & mPA & mIoU  \\
\noalign{\smallskip}\hline\noalign{\smallskip}
		without Domain Adaptation&0.61$\pm$0.05&0.44$\pm$0.05\\
		with Domain Adaptation &\textbf{0.81$\pm$0.07}&\textbf{0.69$\pm$0.1}\\
\noalign{\smallskip}\hline
\end{tabular}
\end{table}

\begin{table}
\label{comparison1}
\centering
\caption{Mean Pixel Accuracy and mean Intersection of Union on the segmentation result on synthetic images.}
\label{comparision1}   
\begin{tabular}{lll}
\hline\noalign{\smallskip}
Method & mPA & mIoU  \\
\noalign{\smallskip}\hline\noalign{\smallskip}
			Segmentation on synthetic images&0.88$\pm$0.04&0.78$\pm$0.07\\
\noalign{\smallskip}\hline
\end{tabular}
\end{table}
\section{Conclusion}
We presented a GAN-based method for unknown objects segmentation mask generation which does not need any labeled data from reality at all. Our method can effectively bridge the sim-to-real gap. We also present a new dataset with point cloud, 6D pose, segmentation, depth and RGB created using the PyBullet, which contains 100k groups of data and provides significantly lots of parameter variations. The details of implementation can be found at \url{ChenEating716.github.io}.
\section*{Acknowledgment}
Funding for this project was provided by Natural Science Foundation of Jiangsu Province (Grant No. BK20180235).

\bibliographystyle{./bibliography/IEEEtran}
\bibliography{./bibliography/IEEEabrv,./bibliography/IEEEexample}

\end{document}